\documentclass{article}

\usepackage[english]{babel}

\usepackage[letterpaper,top=2cm,bottom=2cm,left=3cm,right=3cm,marginparwidth=1.75cm]{geometry}

\usepackage{amsmath}
\usepackage{graphicx}
\usepackage[colorlinks=true, allcolors=blue]{hyperref}

\title{Analyzing Regional Impacts of Climate Change using Natural Language Processing Techniques}
\author{
    Tanwi Mallick \thanks{Mathematics and Computer Science Division, Argonne National Laboratory, Lemont, IL. \texttt{tmallick@anl.gov}} \and
    John Murphy$^{\dagger}$ \thanks{Decision and Infrastructure Sciences Division, Argonne National Laboratory, Lemont, IL. \texttt{\{jtmurphy, dverner, jbergerson, jhutchison, llevy\}@anl.gov}} \and
    Joshua David Bergerson$^{\dagger}$ \and
    Duane R. Verner$^{\dagger}$ \and
    John K Hutchison$^{\dagger}$ \and
    Leslie-Anne Levy$^{\dagger}$
}

\begin{document}
\maketitle

\begin{abstract}

Understanding the multifaceted effects of climate change across diverse geographic locations is crucial for timely adaptation and the development of effective mitigation strategies. As the volume of scientific literature on this topic continues to grow exponentially, manually reviewing these documents has become an immensely challenging task. Utilizing Natural Language Processing (NLP) techniques to analyze this wealth of information presents an efficient and scalable solution.
By gathering extensive amounts of peer-reviewed articles and studies, we can extract and process critical information about the effects of climate change in specific regions. We employ BERT (Bidirectional Encoder Representations from Transformers) for Named Entity Recognition (NER), which enables us to efficiently identify specific geographies within the climate literature. This, in turn, facilitates location-specific analyses.
We conduct region-specific climate trend analyses to pinpoint the predominant themes or concerns related to climate change within a particular area, trace the temporal progression of these identified issues, and evaluate their frequency, severity, and potential development over time. These in-depth examinations of location-specific climate data enable the creation of more customized policy-making, adaptation, and mitigation strategies, addressing each region's unique challenges and providing more effective solutions rooted in data-driven insights.
This approach, founded on a thorough exploration of scientific texts, offers actionable insights to a wide range of stakeholders, from policymakers to engineers to environmentalists. By proactively understanding these impacts, societies are better positioned to prepare, allocate resources wisely, and design tailored strategies to cope with future climate conditions, ensuring a more resilient future for all.
\end{abstract}

\section{Introduction}

Climate change unmistakably stands as one of the most pressing challenges of the 21st century, posing threats to ecosystems, economies, and societies on a worldwide scale. Although its overall ramifications are felt globally, the specific impacts vary significantly across different geographic landscapes. These varied consequences manifest in the form of rising sea levels, intensified coastal erosion, and an increase in extreme weather events, to list a few examples. Therefore, it is crucial to delineate these nuanced, regional impacts to develop localized and tailored interventions.
The scientific community has amassed a substantial amount of research to comprehend the multifaceted effects of climate change. This effort has resulted in an enormous number of peer-reviewed articles, studies, and reports, each offering valuable insights into the impacts of climate change, potential mitigation strategies, and approaches to adaptation. However, manually navigating through this burgeoning corpus of scientific literature to extract region-specific information has turned into a colossal task. The immense volume and complexity of the available data render it nearly impossible for stakeholders, including policymakers, planners, engineers, and scientists, to stay abreast of developments and make well-informed decisions.

Navigating this data-rich environment requires groundbreaking methodologies, and one promising direction is the integration of Natural Language Processing (NLP) tools. As a branch of artificial intelligence, NLP can autonomously analyze and interpret vast amounts of textual data. When applied to climate science, NLP can be a valuable asset, extracting information pertinent to specific geographic areas. 

In recent literature, the application of Natural Language Processing (NLP) in climate change research has been diverse and multifaceted. Sachdeva et al. \cite{sachdeva2022computational} employed machine learning-based NLP techniques to analyze 318 climate action documents from cities committed to net-zero targets. Their research aimed at identifying text patterns indicative of 'ambitious' net-zero goals and conducting a sectoral analysis to discern patterns and trade-offs in climate action themes.
Similarly, Sietsma et al. \cite{sietsma2023next} discussed the challenges and potential of using machine learning, particularly NLP, in tracking and assessing climate change adaptation efforts. This study emphasizes the importance of understanding the capabilities and limitations of machine learning in the context of climate change and adaptation research.
Further contributing to this domain, Krishnan et al. \cite{krishnan2023climatenlp} utilized NLP and ClimateBERT, a pre-trained model, to analyze sentiments in climate change-related tweets. Their objective was to understand public opinions and perceptions of this global challenge. 
Moreover, NLP has proven crucial in analyzing vast weather datasets, aiding in identifying patterns and trends essential for accurate extreme weather forecasts and early warning systems  \cite{tounsi2023systematic, alam2020descriptive, kitazawa2021social, rossi2018early, vayansky2019evaluation, zhou2022victimfinder}. Additionally, its capability to monitor social media has enabled the detection of localized extreme weather events, often overlooked by official reporting \cite{kitazawa2021social, zhou2022victimfinder}. 
These works collectively highlight the expansive and impactful application of NLP in climate change research, from policy analysis and public sentiment assessment to enhancing weather forecasting and recognizing local weather phenomena. The focus has predominantly been on the global effects of climate change, with specific regional implications warranting further exploration in future research.
Duffy et al. \cite{duffy2020leveraging} analyzes the regional distribution of risk and benefit themes in marine aquaculture-news using natural language processing to extract and geocode location data from articles. The findings offer insights for integrating this approach into interdisciplinary research to understand the industry's social and ecological impacts. However, the focus on news media excludes other forms of public discourse and scientific literature that could provide a more comprehensive understanding of the industry's impacts.

To that end, we leverage advanced NLP algorithms, particularly BERT (Bidirectional Encoder Representations from Transformers) \cite{alaparthi2020bidirectional}, and implement Named Entity Recognition (NER) \cite{sharnagat2014named} on a vast collection of climate-focused scientific literature to identify geographic areas and the corresponding climate effects in these regions. Following this identification, we conduct a thorough, geo-centric examination, identifying dominant themes, and evaluating their frequency and severity.

By employing advanced techniques such as BERT’s NER, our objective is to pave the way for precise, geo-specific insights, enhancing our understanding of the potential evolution of climate impacts over time. These detailed analyses of localized climate impacts lay the groundwork for evidence-based policymaking, mitigation strategies, and adaptation planning. 
This paper contributes to the state of practice for analyzing regional impacts of climate change in two major ways: first, we introduce an efficient and scalable framework for analyzing vast scientific corpora; and second, we provide empirical insights that can inform policy formulations and guide resource allocation in the realm of climate adaptation and mitigation. In equipping stakeholders with the tools to understand these regional nuances, we are charting a course toward a more adaptive and resilient global future.


The main contributions of the paper are as follows:
\begin{itemize}

    \item  Development of an advanced NLP pipeline that accurately identifies location information within an unstructured dataset, enabling extraction of geographic-specific climate trends and topics.
    \item Summarization using large language models (LLMs) to efficiently provide a concise understanding of geographic-specific climate change trends.
    \item Development of a vast climate corpus containing over 600,000 documents with document-level tagging of the geographic locations.
    \item Identification of the prevalence of country and state mentions in the climate corpus, providing insights into the regional distribution of climate research.
    \item Integration of all capabilities into a user-friendly tool, streamlining the analytic process for end-user

\end{itemize}

Overall, this work presents a structured and comprehensive approach to analyzing a massive corpus of climate documents, extracting geographically relevant information, and presenting the data in a visual and intelligible manner.

\section{Corpus}
The Semantic Scholar Open Research Corpus (S2ORC) \cite{lo2019s2orc} is a collection of open-access English-language academic papers and scholarly articles. This corpus was created by combining data from various sources and identifying open-access publications. At the time the project team downloaded the corpus, it contained 136 million paper nodes, 76 million abstracts, and 12.7 million full-text papers. 
The corpus covers a breadth of research domains, including medicine, biology, chemistry, engineering, computer science, physics, material science, math, psychology, economics, political science, business, geology, sociology, geography, environmental science, art, history, and philosophy.
Each document in S2ORC is available in a JSON format, and each document has the following entries: paper ID, title, authors, abstract, year, arXiv ID, DOI, journal, field of study, and PDF parse (full PDF availability).

To retrieve documents related to climate change from the S2ORC, we implement semantic similarity search on the corpus \cite{mallick2023analyzing}. As described in the paper \cite{mallick2023analyzing}, we identified 18 climate hazard categories (e.g. extreme cold events, extreme heat events, droughts) through a literature review and established definitions for each category
Using a pre-trained sentence transformer, we generate semantic embeddings for all S2ORC abstracts and each climate category definition. Subsequently, we measure the semantic similarity between the embeddings for abstracts and climate definitions using cosine similarity. This method helps us ascertain if an abstract is associated with any of the 18 climate categories. Through this streamlined approach, we establish a specialized corpus of 604,621 abstracts centered on our climate change hazard categories, derived as a subset from S2ORC.

\section{Methodology}
\begin{figure*}[!ht]
\centering
   \includegraphics[width=\linewidth]{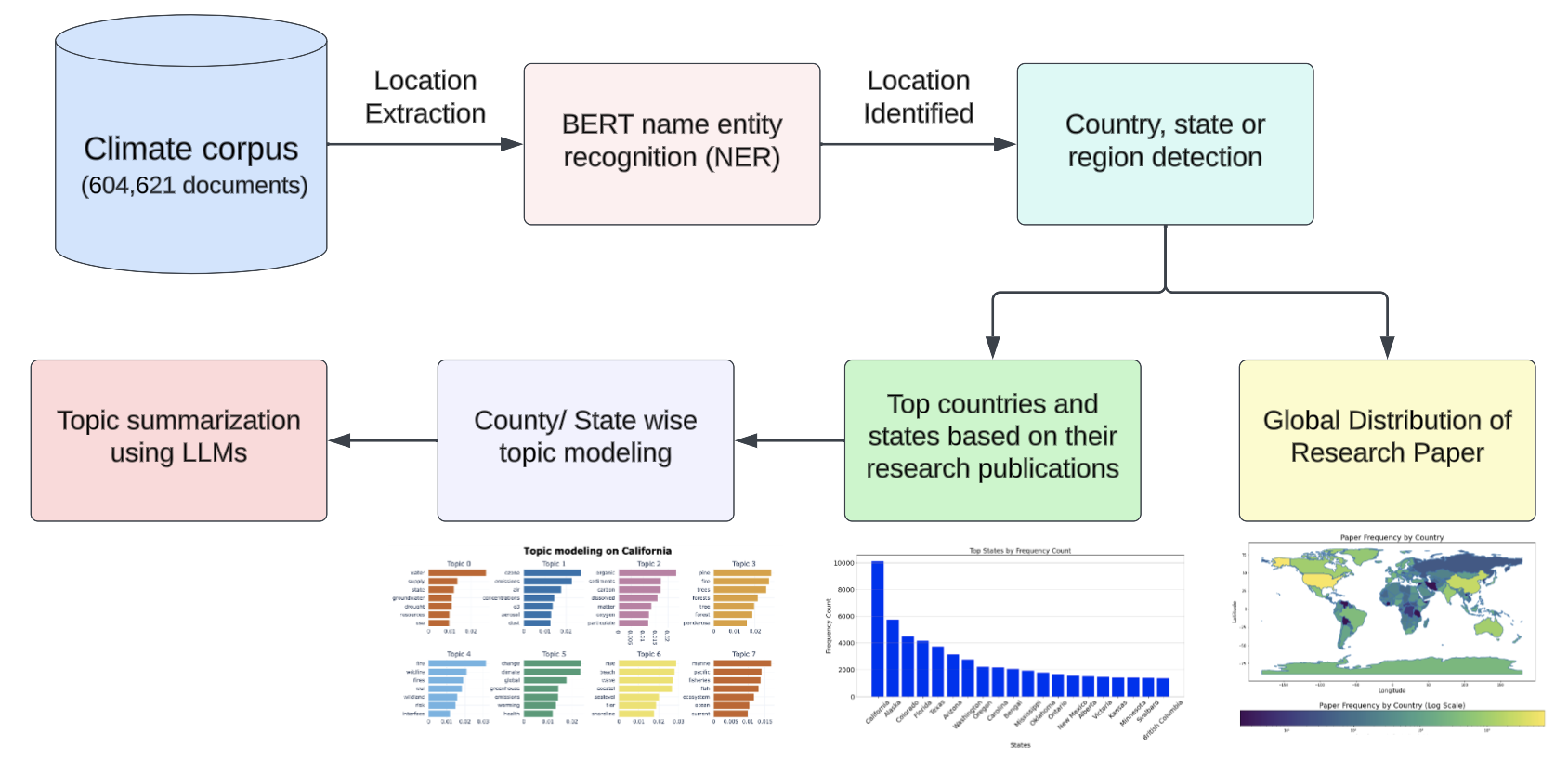}  
\caption{Concept diagram} 
\label{fig_method}
\end{figure*}

In this paper, we study the localized impact of climate change on different geographic locations. The methodology to analyze the impact and elicit specific insights from the large corpus is depicted graphically in Figure \ref{fig_method}.

\subsection{Extracting Geographic Insights from Articles}
In this section, we discuss the process of extracting geographic locations mentioned in the climate change corpus. Leveraging advanced text analysis techniques and location-based data extraction tools, our objective is to identify and catalog every mention of cities, countries, landmarks, and other significant geographical entities. This endeavor not only provides a spatial understanding of the content but also aids in contextualizing the events, stories, or trends covered in the articles. By mapping these mentions, we can gain insights into the regional focus of the content and how different locations play into the broader narrative presented by the collection of articles.
\subsubsection{BERT's Named Entity Recognition}

In our endeavor to glean geographic insights from a climate-focused corpus, we employ BERT's NER \cite{dslim2023bertner}. 
BERT's NER  leverages the power of the BERT architecture to identify and categorize named entities within a text \cite{DBLP:journals/corr/abs-1810-04805}. At its core, BERT is a deeply bidirectional transformer, meaning it reads text from both left to right and right to left, thereby grasping context from both directions. When fine-tuned for NER tasks, each word or subword token is passed through this pre-trained BERT model. The output is a contextualized embedding for each token which captures its semantic meaning in relation to the entire input sequence. This embedding is then passed to a dense layer to predict a NER label for that token. 
Due to BERT's robust understanding of context, its NER capabilities often outperform traditional methods, effectively spotting and classifying entities like names, dates, organizations, and more within a given text. We use BERT's NER to extract location information from the abstracts in our climate-related corpus. BERT's NER identifies four types of entities: locations (LOC), organizations (ORG), persons (PER), and Miscellaneous (MISC). Among these four types of entities, we specifically extract the location information.
Thus, by harnessing BERT's NER, we create a comprehensive geographic map of the climate discourse, pinpointing areas of focus, concern, and discussion within the vast landscape of climate literature.  

\subsubsection{Identify countries, states, and cities}
Following our initial extraction using BERT's NER, we further refine and categorize our geographical data using LocationTagger \cite{locationtagger2023, brunsting2016geotexttagger}. While BERT's NER provides us with a robust set of location-based entities, LocationTagger specializes in distinguishing among countries, states, and cities. This two-step process helps to reduce the noise in the final geographic tags. Once the text has been parsed by BERT's NER for potential geographical mentions, LocationTagger meticulously combs through the results, classifying each mention into its appropriate category. This combination of BERT's contextual understanding and LocationTagger's specialized geographical knowledge yields a comprehensive and granular geographic map of our data, allowing us to identify, with precision, the global spread and regional nuances present within our corpus. 

\subsection{Analyze the tends}
In this section, we discuss different methods for analyzing climate trends specific to geographic locations. First, we identify the countries and states that frequently emerge as central themes. Following this, we discuss topic modeling tailored for these specific geographic locations. Lastly, we focus on topic summarization using LLMs. By leveraging the capabilities of LLMs, we condense extensive information into clear summaries that capture the core essence of each topic.

\subsubsection{States and Countries at the Forefront of Discussion}

After identifying specific countries, states, and cities mentioned in the literature, our subsequent step involves calculating the frequency of each mentioned country, state, or city within the corpus documents. 
The frequency of each country, state, or city mentioned is calculated by determining the number of abstracts in the climate corpus that reference a specific location. This figure is then compared to the total number of abstracts in the climate corpus to establish a relative frequency for each mentioned country, state, or city.
This quantitative approach provides a clear picture of the geographic areas that dominate the discourse on climate change in the literature. By analyzing these frequencies 
, we can discern which states and countries are consistently at the epicenter of the climate discussion, as reflected in the open research corpus. Such insights may suggest strong regional engagement in climate-related dialogues, or may suggest regions expected to experience more significant or rapid impacts driven by climate change. Similarly, these insights may indicate areas where substantial research, interventions, or policy developments are concentrated. While not diminishing these potential understandings, the inherent bias in the underlying corpus should be considered. Our research is based on the Semantic Scholar open research corpus, a vast and comprehensive database, but like all datasets, it is inherently subject to biases. One such bias within the corpus is its exclusion of non-English research texts. Biases can also emerge from various other factors, including the direction of research funding, the prominence of certain research institutions, or regional research priorities. 

\subsubsection{Topic Modeling in Geographical Contexts}
By tagging the documents with specific countries, states, and cities, we can now conduct topic modeling tailored to individual geographic locations. Topic modeling is an effective unsupervised technique for identifying topics present in a set of documents. When applied to individual geographic locations, it helps us extract insights that are localized and directly relevant to individual regions. This enables us to address the unique needs and characteristics of each location.

Latent Dirichlet Allocation (LDA) \cite{blei2003latent} is a widely adopted model that views a document as a collection of words, representing it through a blend of latent topics. However, a limitation of LDA is its inability to consider the semantic relationships between words. To address this, we turned to BerTopic \cite{grootendorst2022bertopic}, which harnesses the BERT embedding technique to capture the semantic nuances within documents.

In its process, BerTopic begins by generating document embeddings using pre-trained models based on BERT. These embeddings then undergo dimensionality reduction through the Uniform Manifold Approximation and Projection for Dimension Reduction \cite{mcinnes2018umap}. Subsequently, the streamlined embeddings are grouped using hierarchical density-based spatial clustering, known as HDBSCAN \cite{mcinnes2017hdbscan}, resulting in clusters of semantically related documents. Each of these clusters signifies a unique topic. To ascertain the significance of words within these clusters, a cluster-based term frequency-inverse document frequency (c-TF-IDF) approach is employed. Unlike traditional TF-IDF which focuses on individual documents, c-TF-IDF evaluates a word's importance within a cluster. 

\subsubsection{Topic summarization using LLMs}

Upon completing the topic modeling process, our next step is to summarize the documents associated with each identified topic. BERTopic provides a list of representative documents for each topic by evaluating their centrality within the cluster. This evaluation is refined using the C-TF-IDF method, ensuring the selection of the most relevant documents.
These documents, being the epitome of their respective topics, provide a comprehensive view of the topic. To make the insights from these documents more accessible and digestible, we employ Large Language Models (LLMs) for text summarization \cite{zhang2023summit, chang2023booookscore}. LLMs, with their advanced NLP capabilities, can distill the essence of documents into concise and coherent summaries. This approach not only preserves the core information and nuances of the original content but also ensures that readers can quickly grasp the primary themes and insights of each topic without delving into the full-length documents. By combining topic modeling with LLM-driven summarization, we offer a streamlined and efficient way to understand and disseminate the wealth of information contained within our dataset.



\section{Results}
In this section, we systematically present the findings derived from our geographic analyses.
The climate data 
have been meticulously processed, evaluated, and visualized to offer clear insights into the core objectives of our research.

\subsection{Top countries and states based on research publications}
\begin{figure*}[!ht]
\centering
   \includegraphics[width=\linewidth]{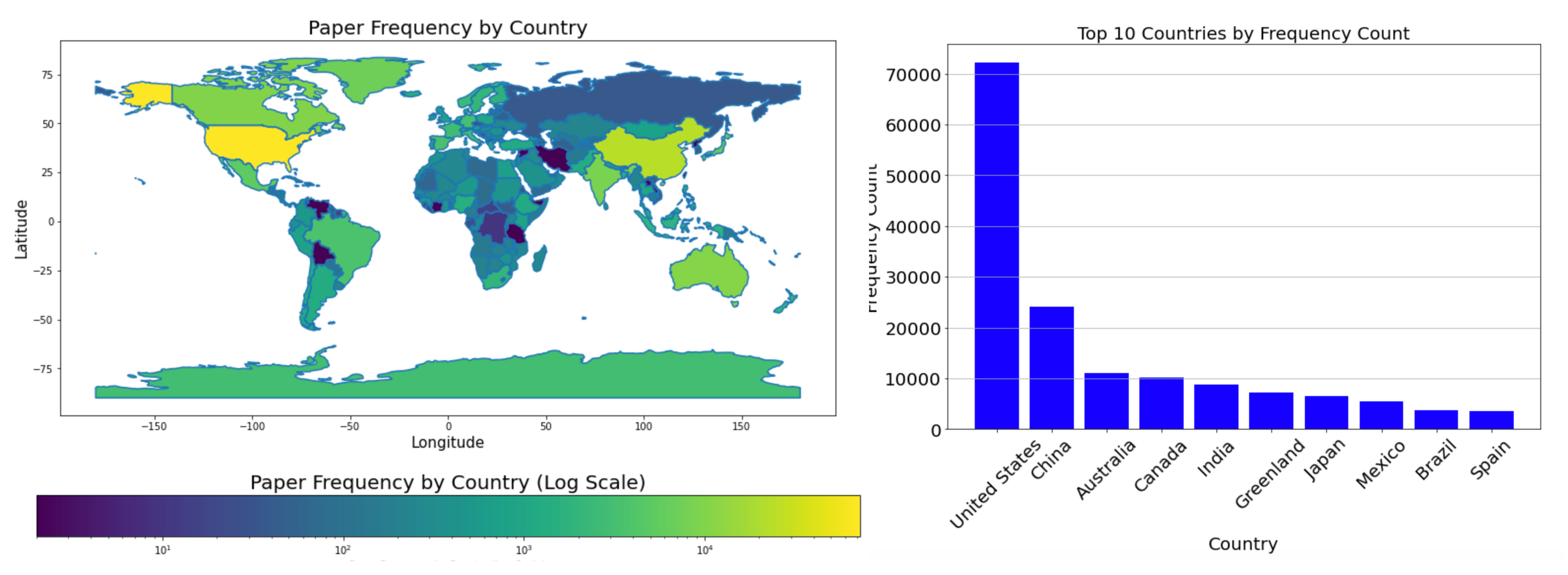}  
\caption{Figure left: Global distribution of research papers. A choropleth map illustrating the frequency of academic papers by country. Lighter shades represent higher frequencies, with a logarithmic scale emphasizing the range of contributions across nations. Figure right: Top 10 countries by mentions in the climate corpus. The bar chart showcases the top 10 countries based on their research paper output. Please note that the data, sourced from the Semantic Scholar's Open Research Corpus, might reflect some inherent biases due to the dataset's nature or the methods employed during its collection and assessment to identify the subset associated with climate change hazards.} 
\label{country_freq}
\end{figure*}

\begin{figure*}[!ht]
\centering
   \includegraphics[width=\linewidth]{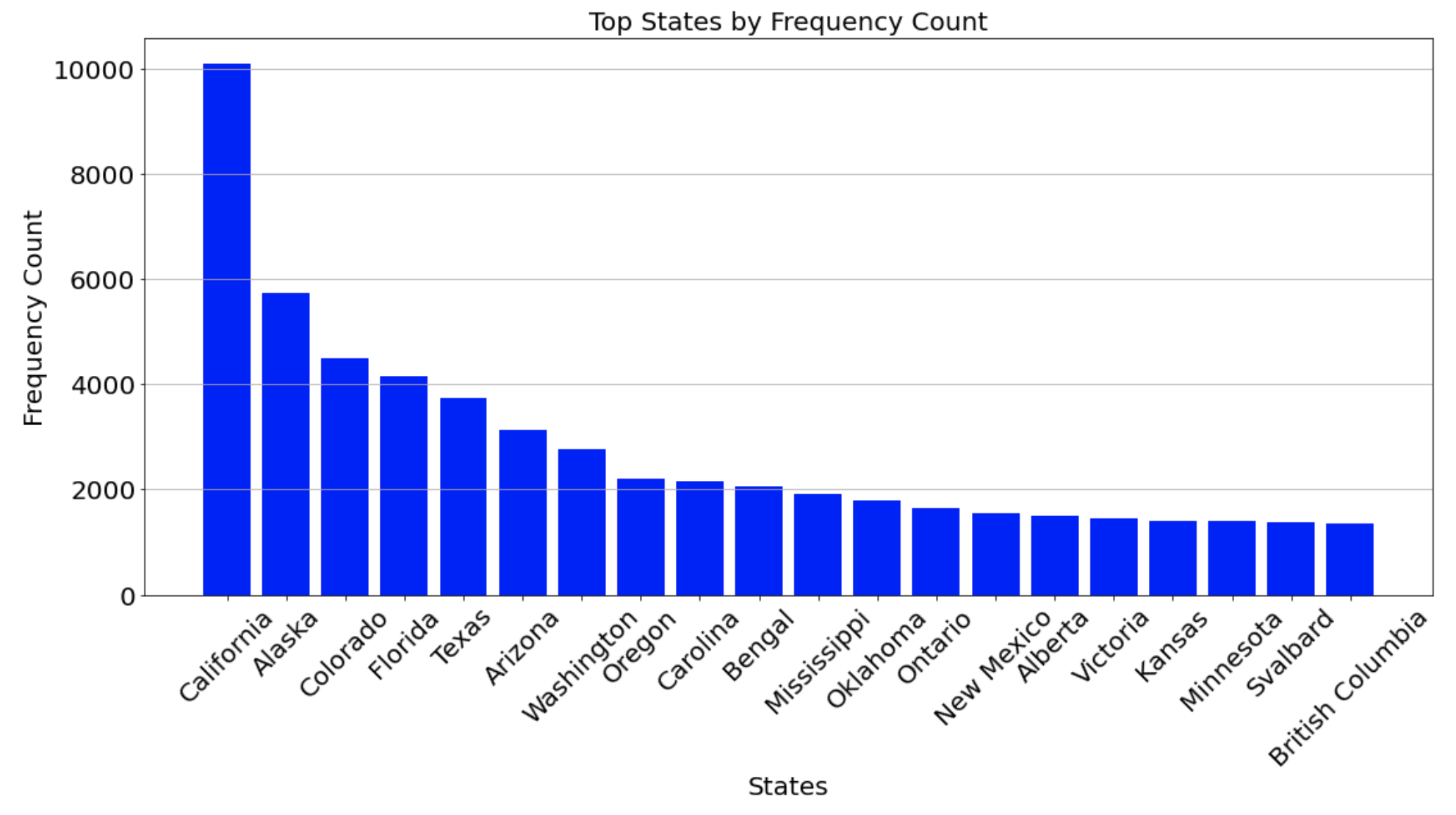}  
\caption{Frequency of documents mentioning states in the climate corpus} 
\label{fig_states}
\end{figure*}

In this section, we count and visualize the distribution of research papers based on their association with specific countries and states. The distribution is calculated by the number of abstracts in the climate corpus that mention a specific country, or state, relative to the total number of abstracts in the climate corpus.
This provides insights into the geographic focus of climate-related studies in our climate corpus, highlighting regions that are frequently addressed in the academic literature. This serves as a reflection of the areas that are perhaps most researched for climate issues.
Figure \ref{country_freq} presents a visualization of a choropleth map that showcases the frequency of papers by country across the world. In the left plot, the color intensity of each country on the map corresponds to the number of papers from that region, with lighter shades indicating a higher frequency and darker shades representing a lower frequency. The color scale, as displayed at the bottom, uses a logarithmic scale, emphasizing the vast range of paper frequencies across different nations in the climate corpus. 
For instance, countries colored yellow have a significantly high number of papers, while those in dark blue have comparatively fewer contributions. The map offers a comprehensive view of global research contributions, highlighting regions with substantial academic and research outputs and those that might be relatively less active in the domain being considered. The presence of the logarithmic scale ensures that even countries with modest contributions are discernible, thus providing a balanced representation. The map acts as a useful tool for understanding the geographic distribution of research output and can be instrumental in guiding policies, funding, and collaboration strategies in the academic and research sectors.
The bar chart in Figure \ref{country_freq} illustrates the frequency of papers mentioning countries for the top 10 countries, ranked by the number of documents mentioning these countries. Clearly dominating the chart, the United States boasts a count that surpasses 70,000, indicating a prodigious volume of research output. Following the U.S., China has also made a significant contribution, albeit at a frequency count around half of that of the U.S. As we move further to the right, countries such as Australia, Canada, and India demonstrate commendable contributions, with their counts ranging between 10,000 and 30,000. Greenland, Japan, Mexico, Brazil, and Spain complete the top 10 list, with their counts decreasing progressively, yet still notable. The declining height of the bars from left to right visualizes the drop in frequency counts, offering a comparative perspective on the geographic focus of the research landscape. This chart provides valuable insights into the global distribution of discussions on climate change in research papers and underscores the pivotal role of certain countries in driving academic and research advancements in the domain. We acknowledge the potential geographic bias in our dataset and assessment, such as the exclusion of non-English publications, which may affect the expected geographic distribution of climate research within our climate corpus.

Figure \ref{fig_states} showcases the frequency of papers in the climate corpus mentioning states and regions for the top 20 such geographies. California stands out with the highest frequency, indicating a significant volume of research papers associated with this region. This could be attributed to California's diverse climate, its susceptibility to climate-related challenges, or the presence of major research institutions. Following California, there is a noticeable decline in frequency, with states like Alaska, Colorado, and Florida having a moderate representation. As we move further right in the figure, the frequencies become relatively consistent, suggesting a more evenly distributed volume of research across these states. Interestingly, the chart also features locations like Bengal, Victoria, and Svalbard, from other countries such as India, Australia, and Norway. This inclusion emphasizes the global nature of climate studies and the importance of various regions in contributing to the discourse. Moreover, the inclusion of multiple states or regions from the Arctic, such as Alaska and Svalbard, highlights the crucial role the Arctic plays in climate research. The Arctic is often considered the proverbial "canary in the coal mine"  for global warming due to the rapid and pronounced changes observed in this region \cite{stephenson2018confronting, eicken2013arctic}. 
As a part of the U.S., Alaska has been a focal point for numerous climate studies given its vast glaciers, permafrost landscapes, and unique ecosystems. Svalbard, on the other hand, is a group of islands in the High Arctic known for its polar climate and its sensitive response to global temperature changes. Svalbard is also home to the Svalbard Global Seed Vault, a vitally important global resource for preserving biodiversity. The prominence of these Arctic regions in the chart underscores the urgency and significance of understanding climate dynamics in these areas. Research focused on these locations offers invaluable insights into the broader implications of climate change, serving as early indicators for global environmental shifts.


\subsection{Topic Modeling}

\begin{figure*}[!ht]
\centering
   \includegraphics[width=\linewidth]{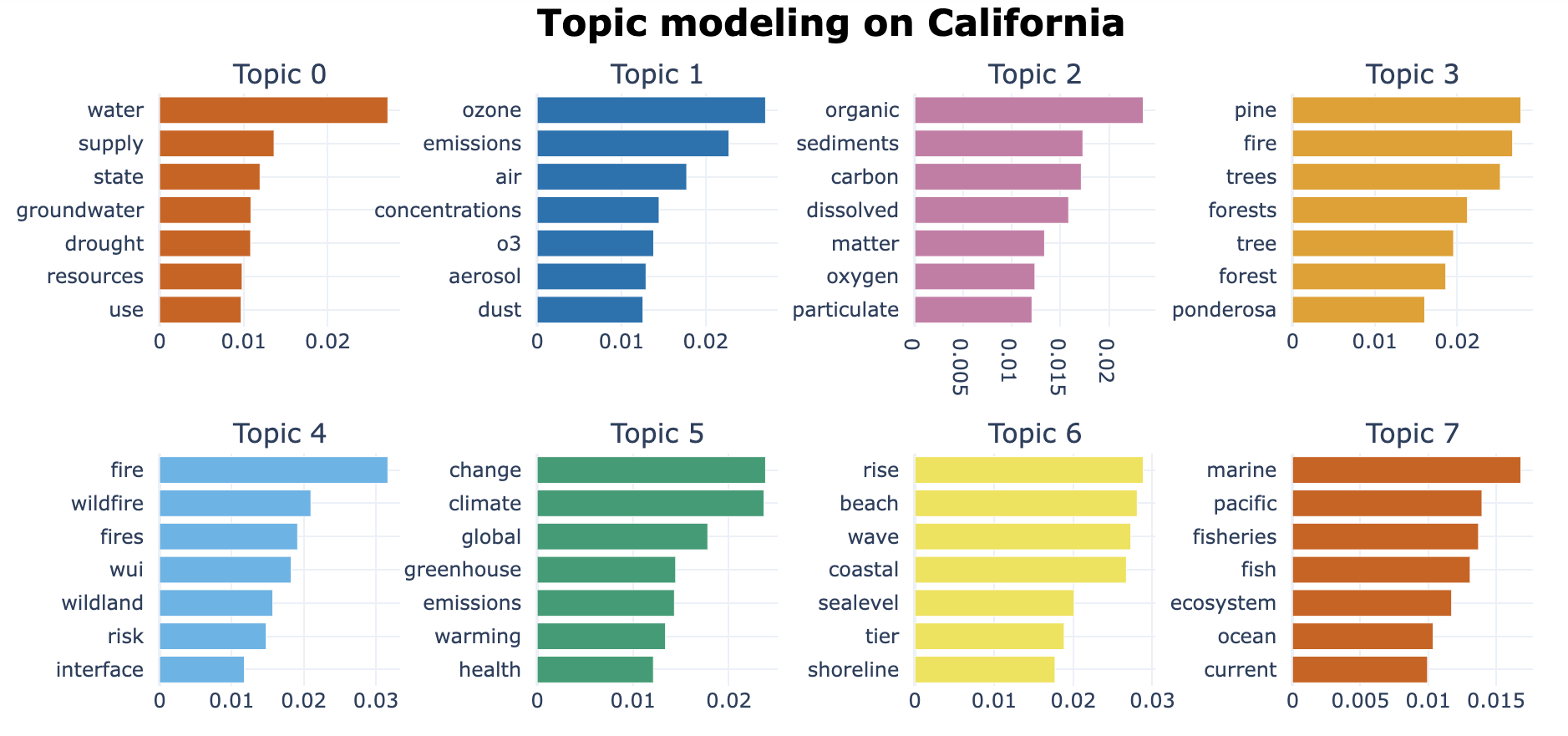}  
\caption{Topic distributions for climate research papers focusing on California are uncovering dominant themes and areas of study.} 
\label{fig_tm_cali}
\end{figure*}

We carried out topic modeling on geographically focused documents to gain insights into the climate dynamics of various regions. Figure \ref{fig_tm_cali} showcases the results of topic modeling conducted on climate research documents related to California, which emerged as a state with frequent mentions in the climate corpus. Each bar chart represents a distinct topic, labeled from Topic 0 to Topic 7, and provides a visualization of the keywords associated with each topic.  The size of the bars in each bar chart corresponds to the relative frequency of each keyword within its respective topic. 
Examining Figure \ref{fig_tm_cali}, Topic 0 largely pertains to water resource management with keywords like water, supply, resources, drought, and state. In contrast, Topic 4 talks about wildfires, highlighting terms such as fire, wildfire, and risk. Topic 5 offers insights into broader climate change dynamics, featuring words like change, global, and warming. Similarly, each of the other topics captures a unique aspect of California's environmental and climate research landscape, from marine ecosystems to sea-level rise. Furthermore, a deeper dive into these topics sheds light on the nuanced narratives underpinning California's environmental challenges driven by climate change and population dynamics. The state's urban centers, grappling with recurring droughts, are on a continuous quest for innovative water conservation measures \cite{tanaka2006climate, aghakouchak2015water}. 
Simultaneously, the symbiotic relationship between water and energy emerges as a significant theme, emphasizing the dual role of water in energy production and the energy-intensive nature of water management.
The economic dimension is equally compelling. 

\begin{figure*}[!ht]
\centering
   \includegraphics[width=\linewidth]{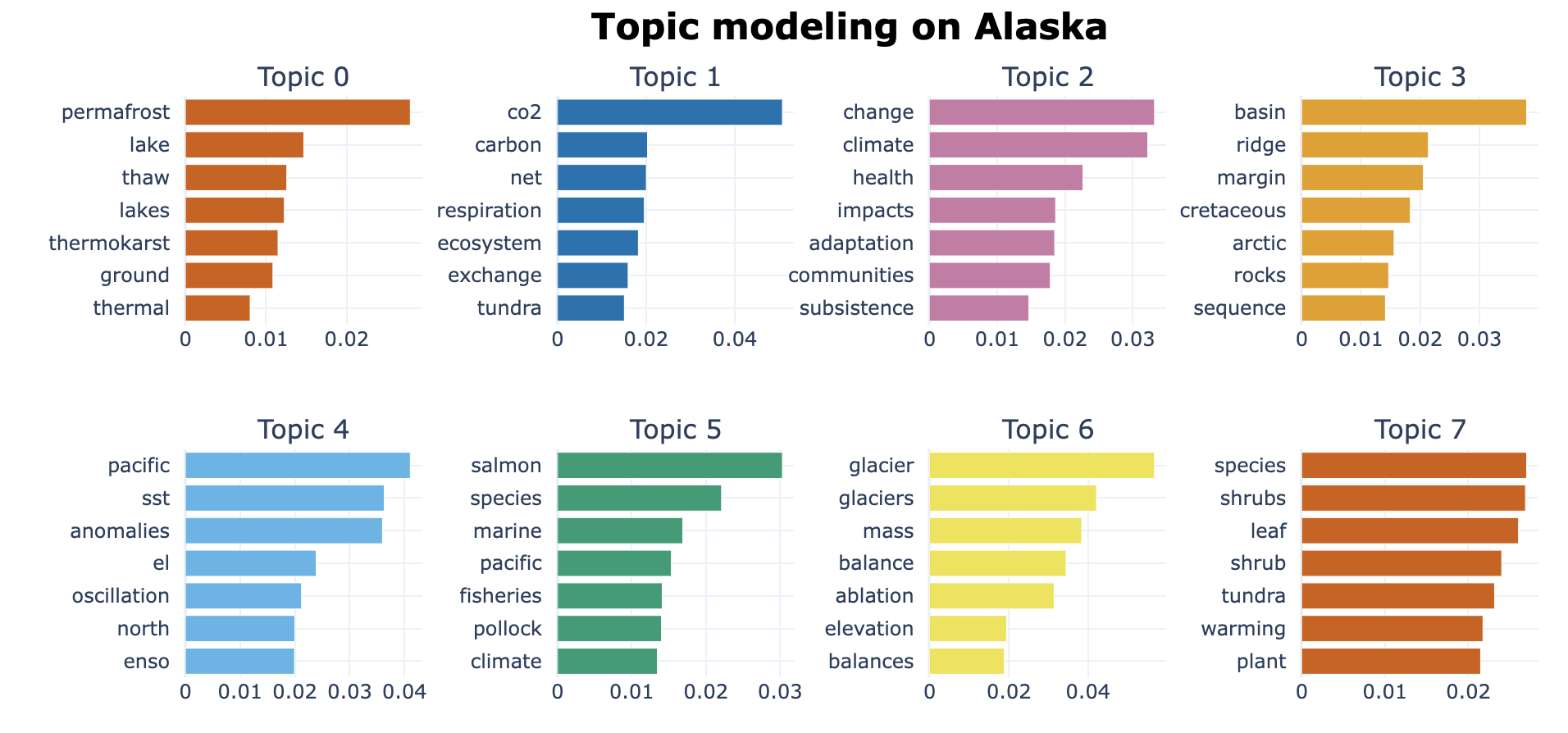}  
\caption{Topic distributions for climate research papers focusing on Alaska are uncovering dominant themes and areas of study.} 
\label{fig_tm_alaska}
\end{figure*}

Figure \ref{fig_tm_alaska} showcases the results of topic modeling conducted on climate-related documents specific to Alaska. Each bar chart, labeled from Topic 0 to Topic 7, represents a distinct area of focus. Topic 0 predominantly highlights issues related to permafrost thaw, with terms like permafrost, lake, thaw, and thermokarst hinting at the melting and changing landscape. Topic 1 delves into the carbon cycle in Alaska's unique ecosystems, with terms such as co2, carbon, and respiration coming to the fore. Topic 2 touches upon broader climate change impacts and the resultant challenges and adaptations for local communities. On the other hand, Topic 3 seems to focus on geological aspects, including terms like basin and rocks. The influence of the Pacific Ocean and its associated climatic phenomena, like El Niño, is evident in Topic 4. Topic 5 emphasizes Alaska's marine ecosystem, with a specific focus on species like salmon and their relationship with the climate. The presence of glaciers and related terms in Topic 6 indicates discussions around glacial melt and its implications. Lastly, Topic 7 underscores the vegetation changes in the region, particularly in the tundra, highlighting terms like shrubs, warming, and plants. Overall, the Arctic holds immense significance in the global ecosystem. As a barometer for climate change, its rapidly melting ice caps and glaciers serve as early indicators of global warming's impact. The region plays a crucial role in global weather patterns and sea level regulation. Moreover, the Arctic is rich in biodiversity and is home to unique flora and fauna, making it a vital habitat for many species. Additionally, its vast reserves of untapped natural resources have economic implications, while the cultural heritage of its indigenous communities offers invaluable insights into human adaptability and resilience.

Overall, through topic modeling, we've distilled multifaceted and interconnected themes that define the environmental and climatic landscape of California and Alaska. This synthesis not only offers a snapshot of the current research landscape but also underscores the myriad challenges and opportunities that lie ahead for each state. The insights gained from conducting topic modeling on climate research papers discussing each state demonstrate the potential benefits of applying this method to other states. Like California and Alaska, we can perform topic modeling to understand the environmental and climatic discourses of other regions. Each state, with its unique geographic, ecological, and socio-economic attributes, will undoubtedly present its own set of dominant themes and concerns. By extending this analytical approach to other states, we can construct a comprehensive, state-by-state tapestry of environmental research trends across the nation. This would not only offer localized insights but also enable comparative analyses, highlighting shared challenges, and distinctive state-specific issues. 

\subsection{Topic Summarization using ChatGPT}

The final component of our developed approach utilizes LLM functionality to produce succinct and coherent summaries of each topic identified through topic modeling. 
After identifying the most representative documents for each topic, we feed these documents to an LLM to develop short summaries for each topic. For this step, we selected to use ChatGPT as the LLM given its existing user interface reducing the barrier to entry and its widespread use across industries. This method for developing topic summaries allows us to capture the essence of vast amounts of data, providing researchers, policymakers, and the general public with a clear snapshot of the primary themes and findings within a topic. Instead of sifting through pages of dense academic content, users can quickly grasp the central messages and nuances, facilitating more informed decision-making and discussions. Here is the 50-word summary of the top 3 topics previously identified for California (Figure \ref{fig_tm_cali}):

\begin{itemize}
    \item {\bf Topic 0}: {\em ``Water conservation efforts in California's urban households during recent droughts are analyzed using survey data from Los Angeles and San Francisco. Energy and water connections in California are highlighted, emphasizing the complexities of their economic relationship. With limited storage solutions, water agencies are exploring contractual mechanisms to manage supply risks. Experimental economics is used to study the impact of dry-year options in water markets, finding that options increase realized gains in trade. This research aids policymakers in drought preparedness.''}
    \item {\bf Topic 1}: {\em ``The studies investigate various aspects of atmospheric emissions and aerosols. The first focuses on seasonal methane emissions in Northern California, revealing significant underestimates in prior emission inventories. The second examines aerosol composition in Mira Loma, CA and the Colorado Front Range, linking air quality degradation to specific aerosol types and meteorological phenomena. The third study highlights the impact of distant emission sources on observed haze events near the Grand Canyon.''}
    \item {\bf Topic 2}: {\em ``Research into sediment diagenesis has enhanced our comprehension of organic matter remineralization and preservation in seafloors. Studies indicate that input plays a more significant role than preservation factors in organic carbon burial. Meanwhile, a 6-year time series on organic matter in the Northeast Pacific Ocean suggests regional meteorological events influence particle properties in deep-sea locations. Lastly, the discovery of microbes in the Guaymas Basin that anaerobically degrade gaseous alkanes underscores the biofiltering role of these organisms and introduces potential new mechanisms of alkane oxidation.''}
\end{itemize}

\section{Database and Tool}
The suite of methodological components described earlier in this paper, as well as additional methodological components described in \cite{mallick2023analyzing}, are integrated into a decision-support tool called the Community and Infrastructure Adaptation to Climate Change (CIACC) tool. The tool is facilitated by an underlying back-end database that stores the underlying corpus and the additional data about the corpus that we have generated through the processes described above and in \cite{mallick2023analyzing}. The hardware and software architecture that we use, and the functions each components serves, are documented in the subsequent sections.

\subsection{Solr Database}

A Solr database, available at \cite{ApacheSolr} 
, is hosted on a 128GB iMac and holds detailed document-level data. This includes the complete collection of abstracts utilized in our study. Additionally, the database contains all the labels designated to these documents. These labels encompass the identified geographic entities, and topics pinpointed by our topic modeling (along with their respective scores). 


While Solr is one of many non-SQL databases available, the project team selected to use a Solr database due to its performance and integration with the remainder of the components of our approach. In future work, a more thorough evaluation of other technologies may be required, but none of the work presented here tested Solr's performance or scalability.

\subsection{MySQL Database}
A MySQL database \cite{MySQL} 
was used to store metadata about the groups, topics, and geographic entities identified in the Solr database. This database worked in conjunction with the Solr database, allowing the project team to leverage the capabilities of both database structures. For example, when a topic model was created, the MySQL database would store the (user-assigned) name of the model and the list of the top words for each topic created by the model. These were assigned arbitrary, numeric primary keys. The Solr database was then updated, so that each document in the corpus that was related to the topic was marked as such. The advantage of this structure is that the MySQL structured database could contain the relatively limited amount of metadata, while the Solr database could be restricted to storing document-level data. The result is that a topic model with 10 topics and a small collection of relevant words would amount to perhaps 50 records in the MySQL database, while the Solr database marked thousands of documents with the scores related to those topics. 

\subsection{BERTopic Modeling Platform}
BERTopic modeling was performed on a custom-built PC running the Linux operating system Ubuntu. BERTopic requires a Python environment, and a custom environment was established for performing the BERTopic modeling.

\subsection{Apache Tomcat Web Server}
The user interface was provided via an Apache Tomcat web server and collection of .jsp interfaces \cite{ApacheTomcat}. 
Together, the server and underlying Java code provide the following capabilities to users:
\begin{enumerate}
\item Access to all user-created document groups, and allowing the review and creation of such groups (and sets of groups termed 'Group Sets'). 
\item Access to all topic models, and allowing the review and creation of new topic models, including retrieving documents marked as related to each topic.
\item Access to all geographic entities, and allowing the review of geographic entities and of the documents associated with each entity.
\item Additional graphical summarization functionality; for example, plotting the intersections of sets of documents and of the prominence of specific topics within given topic models
\item Ability to directly query the Solr database
\end{enumerate}

The Tomcat server and underlying code were responsible for maintaining the integrity of the MySQL database, dispatching and retrieving topic models, and updating and retrieving documents from the Solr database. For example, a user could use the interface to query the Solr database for all documents referencing a specific geographic entity. In this scenario, the interface would retrieve the count of documents and allow users to review the list of documents or read individual documents, as needed.  The interface then allowed the user to request a topic model across all of these documents. The underlying code would update the MySQL database to reflect the creation of a new topic model, copy the documents needed to the BERTopic Modeling Platform, initiate the topic modeling process, and periodically check its progress. Upon completion, the system retrieved the topic model, added the new topics into the MySQL database, and stored the documents' scores in the Solr database, making them available for retrieval and review. Once stored, a graphical summary of the model could be reviewed, the full list of documents in each topic retrieved, or individual documents inspected.


\begin{figure*}[!ht]
\centering
   \includegraphics[width=\linewidth]{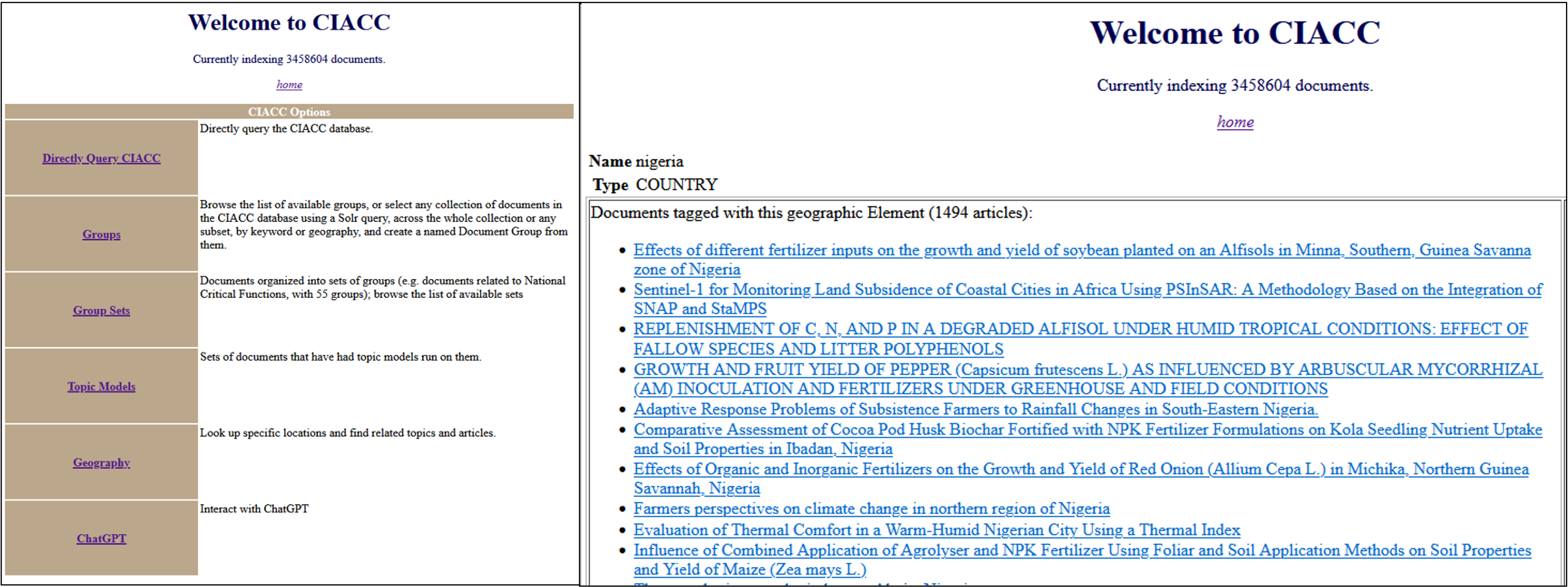}  
\caption{Overview of the CIACC Interface. A Comprehensive Platform for Document Exploration and Research Analysis, featuring Direct Querying, Thematic Groupings, Geographic Tagging, and Advanced AI Assistance with ChatGPT.} 
\label{fig_tool}
\end{figure*}

Figure \ref{fig_tool} presents a user interface for the CIACC database platform, an expansive system that currently indexes a remarkable total of 3,458,604 documents and 600k climate-related documents. The CIACC platform offers a meticulously structured interface that empowers users to delve into a vast database of documents with precision and ease. At its core, the Directly Query CIACC feature enables users to enter specific search queries tailored to their research needs. For those looking for thematic guidance, the Groups option curates available document clusters, hinting at a systematic categorization based on distinct subjects or themes. For users aiming to discern overarching trends or dominant themes in the database, the Topic Models offers a window into the prevalent narratives and subjects. Meanwhile, the Geography tool refines searches on a territorial basis, allowing research to be geographically contextualized. Lastly, the inclusion of ChatGPT assists users in their exploration and understanding of the database's expansive content. On the right side of Figure \ref{fig_tool}, we see a highlighted case study of Nigeria labeled as a COUNTRY. Under this label, there's a list of documents tagged with this geographic element, totaling 1,494 articles. This section showcases specific research titles related to Nigeria, giving an impression of the diverse range of topics that the CIACC database covers. From agricultural studies on soybean growth and fertilizer usage to investigations on land subsidence in African coastal cities, the listed documents reflect a wide range of academic and research-oriented content related to Nigeria.

\section{Conclusion}

In summary, our research offers an innovative approach to analyzing climate-focused documents, utilizing an extensive corpus of over 600,000 documents. The adoption of BERT NER facilitated the efficient extraction of location-specific information, which subsequently allowed for precise detection of countries, states, and regions within documents using LocationTagger. We also ventured into topic modeling at regional levels, illuminating intricate climate-related trends that vary geographically. Our method's strength is further underscored by our capability to summarize these topics using LLMs. Additionally, by evaluating the volume of research publications, we've highlighted the countries and states most frequently discussed in climate research. The resulting global distribution of research papers not only mirrors the current research landscape but also underscores regions leading climate research endeavors. This comprehensive approach promises to guide future climate studies and policy-making by providing an in-depth, location-centric perspective on prevalent climate issues. Furthermore, to ensure the accessibility and practical applicability of our research, we developed an interactive tool based on our developed methodological pipeline. This tool empowers users to navigate, visualize, and understand location-specific climate trends seamlessly. Its user-friendly interface simplifies complex data, making our findings more digestible for both researchers and the general public. The integration of this tool not only augments the usability of our methodology but also sets the stage for future expansions and iterative improvements as new data emerges.

\section*{Acknowledgments}
This material is based in part upon work supported by the Laboratory Directed Research and Development (LDRD) program, Argonne National Laboratory. 
Additionally, this research used resources from the Argonne Leadership Computing Facility, which is a DOE Office of Science User Facility under contract DE-AC02-06CH11357. 

\section*{Government license}
The submitted manuscript has been created by UChicago Argonne, LLC, Operator of Argonne National Laboratory ("Argonne"). Argonne, a U.S. Department of Energy Office of Science laboratory, is operated under Contract No. DE-AC02-06CH11357. The U.S. Government retains for itself, and others acting on its behalf, a paid-up nonexclusive, irrevocable worldwide license in said article to reproduce, prepare derivative works, distribute copies to the public, and perform publicly and display publicly, by or on behalf of the Government.  The Department of Energy will provide public access to these results of federally sponsored research in accordance with the DOE Public Access Plan. http://energy.gov/downloads/doe-public-access-plan.

\bibliographystyle{plain}
\bibliography{sample}

\end{document}